\documentclass{article}

\usepackage[final]{neurips_2023}

\usepackage[utf8]{inputenc} 
\usepackage[T1]{fontenc}    
\usepackage{hyperref}       
\usepackage{url}            
\usepackage{booktabs}       
\usepackage{amsfonts}       
\usepackage{nicefrac}       
\usepackage{microtype}      
\usepackage{xcolor}         
\usepackage{amsmath}
\usepackage{graphicx}
\usepackage{multirow}
\usepackage[capitalise]{cleveref}

\usepackage{tikz}
\usepackage{subcaption}
\usepackage{soul} 
\usepackage{wasysym}

\definecolor{darkred}{rgb}{1,0.5,0.5}
\definecolor{bluehighlight}{rgb}{0.8,0.8,1}
\definecolor{lightblue}{RGB}{173,216,230}
\definecolor{lightred}{RGB}{255,182,193}
\definecolor{lightpurple}{RGB}{216,191,216}
\definecolor{lightgray}{RGB}{230,230,230}
\definecolor{darkerlightblue}{RGB}{73,116,130}
\definecolor{darkerlightred}{RGB}{155,82,93}
\definecolor{darkerlightpurple}{RGB}{116,91,116}
\definecolor{darkerlightgray}{RGB}{130,130,130}

\sethlcolor{bluehighlight}
\newcommand{\hlred}[1]{{\sethlcolor{lightred}\hl{#1}}}
\newcommand{\hlblue}[1]{{\sethlcolor{lightblue}\hl{#1}}}
\newcommand{\hlgray}[1]{{\sethlcolor{lightgray}\hl{#1}}}

\usepackage{float}

\usepackage[backgroundcolor=white, textsize=small, textwidth=35mm]{todonotes}

\newcommand{\charlie}[1]{}
\newcommand{\alex}[1]{}
\newcommand{\elias}[1]{}
\newcommand{\diogo}[1]{}
\newcommand{\victor}[1]{}
\newcommand{\bogdan}[1]{}
\newcommand{\edoardo}[1]{}

\newcommand{\charlieC}[1]{}
\newcommand{\alexC}[1]{}
\newcommand{\eliasC}[1]{}
\newcommand{\diogoC}[1]{}
\newcommand{\victorC}[1]{}
\newcommand{\bogdanC}[1]{}
\newcommand{\edoardoC}[1]{}

\title{Reinforcement Learning Fine-tuning of Language Models is Biased Towards More Extractable Features}

\author{
Diogo Cruz\textsuperscript{1}\quad
Edoardo Pona\textsuperscript{1}\\
\textbf{Alex Holness-Tofts}\textsuperscript{1}\quad
\textbf{Elias Schmied}\textsuperscript{1}\quad
\textbf{Victor Abia Alonso}\textsuperscript{1}\\
\textbf{Charlie Griffin}\textsuperscript{1}\quad
\textbf{Bogdan-Ionut Cirstea}\textsuperscript{1} \\
\textsuperscript{1}AI Safety Hub Labs \\
}

\begin{document}

\maketitle

\begin{abstract}
  Many capable large language models (LLMs) are developed via self-supervised pre-training followed by a reinforcement-learning fine-tuning phase, often based on human or AI feedback. During this stage, models may be guided by their inductive biases to rely on simpler features which may be easier to extract, at a cost to robustness and generalisation. 
We investigate whether principles governing inductive biases in the supervised fine-tuning of LLMs also apply when the fine-tuning process uses reinforcement learning. Following \citet{lovering_predicting_2021}, we test two hypotheses: that features more \emph{extractable} after pre-training are more likely to be utilised by the final policy, and that the evidence for/against a feature predicts whether it will be utilised. 
Through controlled experiments on synthetic and natural language tasks, we find statistically significant correlations which constitute strong evidence for these hypotheses.
\end{abstract}

\section{Introduction}

Most capable large language models (LLMs) are developed using self-supervised pre-training, where they learn representations of various features, followed by a reinforcement-learning (RL) fine-tuning phase, during which they learn to utilise these features to perform a specific task according to human preferences \citep{christiano_deep_nodate,ziegler_fine-tuning_2020,stiennon_learning_2020,ouyang_training_2022}. The reward signal provided during the fine-tuning process under-determines the behaviour of the learned policy on data outside the training distribution \citep{damour_underspecification_2022,jayawardana_impact_2022}.

\citep{lovering_predicting_2021} demonstrated that LLM-supervised fine-tuning exhibits the following inductive bias: fine-tuned models are more likely to rely on features that are more extractable after pre-training, even if these features have less predictive power. 
We examine whether this inductive bias also holds for LLMs fine-tuned via RL and when the reward is provided by another learned model \citep{ziegler_fine-tuning_2020,stiennon_learning_2020}. 
Specifically, our contribution is to test the following hypotheses about the policy of the fine-tuned model:

\textbf{Extractability hypothesis}: features which score higher in extractability for the pre-trained model are relied upon more by the RL fine-tuned model (policy).

\textbf{Evidence hypothesis}: the more evidence there is for/against a feature during RL fine-tuning, the more likely the model learns a policy that relies on that feature to get a high reward (policy).

After providing key terminology in \cref{sec:setup_and_terminology}, we explain the experimental setup (\cref{sec:experimental_setup}) and present evidence in support of the extractability hypothesis in \cref{sec:naturalistic_data}. Our key result is \cref{fig:evidence_sentiment}. We then discuss our results (\cref{sec:discussion}) and contrast them against existing work (\cref{sec:related_work}).

\section{Background}\label{sec:setup_and_terminology}

We modify the supervised fine-tuning setup of \citet{lovering_predicting_2021} so that - instead of a binary classification task - the pre-trained language model receives a reward signal from a reward model trained using human labels \citep{ziegler_fine-tuning_2020,stiennon_learning_2020,christiano_supervising_2018,bai_constitutional_2022}. We adopt the definitions of \textbf{evidence} and \textbf{extractability} from \citep{lovering_predicting_2021}, while adapting the definitions of \textbf{target} and \textbf{spurious} features to suit a reinforcement learning setting.

The reinforcement learning problems we consider vary in the reward functions but share common state and action spaces.
Let $X$ be the set of all possible natural language prompts for the task of interest. 
Consider the setup where an LLM takes a prompt text $x\in X$ and produces a response $y$. The initial state distribution is made by sampling from a training dataset.
During the RL fine-tuning process, there may be a \textbf{target feature} $t: X \mapsto \{0,1\}$ in the training data whose presence ($t(x)=1$) or absence ($t(x)=0$) determines the goal for that prompt.
That is, the reward function scoring the prompt-response pair $(x,y)$ can be described as
\begin{equation}
    R(x,y) = \begin{cases}
        R_0(x,y), &\text{if } t(x) = 0\\
        R_1(x,y), &\text{if } t(x) = 1 \\
    \end{cases}
\end{equation}
where $R_0$ and $R_1$ do not depend on $t$. Along with the target feature, there may be \textbf{spurious features} $s$ in the training prompts, whose presence and absence correlate with that of the target feature. The LLM may then learn to get high reward by relying on $s$ instead of $t$ during RL fine-tuning. Our study considers the simplified scenario where only one spurious feature may be present in the prompt.

\section{Experimental setup}\label{sec:experimental_setup}

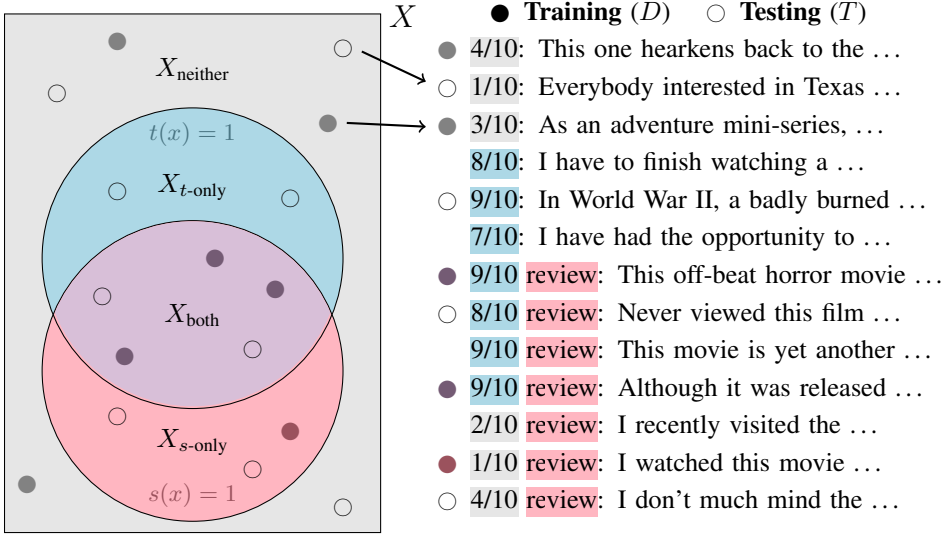
\begin{figure}[!htpb]
\begin{center}
\begin{tikzpicture}
    \tikzstyle{prompt}=[text width=7.5cm]
    
    \node at (7.,3.5) [prompt] {\qquad\newmoon\, \textbf{Training} ($D$) \quad \fullmoon\, \textbf{Testing} ($T$)};
    \node at (7.,3.) [prompt] {\textcolor{darkerlightgray}{\newmoon}\, \hlgray{4/10}: This one hearkens back to the \ldots};
    \node at (7.,2.5) [prompt] {\fullmoon\, \hlgray{1/10}: Everybody interested in Texas \ldots};
    \node at (7.,2.) [prompt] {\textcolor{darkerlightgray}{\newmoon}\, \hlgray{3/10}: As an adventure mini-series, \ldots};
    \node at (7.,1.5) [prompt] {\phantom{\newmoon\,} \hlblue{8/10}: I have to finish watching a \ldots};
    \node at (7.,1.) [prompt] {\fullmoon\, \hlblue{9/10}: In World War II, a badly burned \ldots};
    \node at (7.,0.5) [prompt] {\phantom{\newmoon\,} \hlblue{7/10}: I have had the opportunity to \ldots};
    \node at (7.,0.) [prompt] {\textcolor{darkerlightpurple}{\newmoon}\, \hlblue{9/10} \hlred{review}: This off-beat horror movie \ldots};
    \node at (7.,-0.5) [prompt] {\fullmoon\, \hlblue{8/10} \hlred{review}: Never viewed this film \ldots};
    \node at (7.,-1.) [prompt] {\phantom{\newmoon\,} \hlblue{9/10} \hlred{review}: This movie is yet another \ldots};
    \node at (7.,-1.5) [prompt] {\textcolor{darkerlightpurple}{\newmoon}\, \hlblue{9/10} \hlred{review}: Although it was released \ldots};
    \node at (7.,-2.) [prompt] {\phantom{\newmoon\,} \hlgray{2/10} \hlred{review}: I recently visited the \ldots};
    \node at (7.,-2.5) [prompt] {\textcolor{darkerlightred}{\newmoon}\, \hlgray{1/10} \hlred{review}: I watched this movie \ldots};
    \node at (7.,-3.) [prompt] {\fullmoon\, \hlgray{4/10} \hlred{review}: I don't much mind the \ldots};

    \fill[lightgray] (-2.5,-3.4) rectangle (2.5,3.5);
    \draw (-2.5,-3.4) rectangle (2.5,3.5);
    \node[above] at (0,2.5) {$X_{\text{neither}}$};
    \node[above] at (2.8,3.2) {\large $X$};
    \node[above] at (-1.8,2.2) {\fullmoon};
    \node[above] at (2,2.8) {\fullmoon};
    \node[above] at (2,-3.3) {\fullmoon};
    \node[above] at (-1,2.9) {\textcolor{darkerlightgray}{\newmoon}};
    \node[above] at (1.8,1.8) {\textcolor{darkerlightgray}{\newmoon}};
    \node[above] at (-2.2,-3.) {\textcolor{darkerlightgray}{\newmoon}};
    \draw[->, thick] (2.25,3.) -- (3.1,2.6);
    \draw[->, thick] (2.05,2.05) -- (3.1,2);

    \fill[lightblue] (0,0.25) circle [radius=2.];
    \node at (0,1.2) {$X_{t\text{-only}}$};
    \node at (0,1.9) {\textcolor{darkerlightblue}{\small$t(x)=1$}};
    \node[above] at (-1,0.9) {\fullmoon};
    \node[above] at (1.3,0.8) {\fullmoon};

    \fill[lightred] (0,-1.25) circle [radius=2.];
    \node at (0,-2.2) {$X_{s\text{-only}}$};
    \node at (0,-2.9) {\textcolor{darkerlightred}{\small$s(x)=1$}};
    \node[above] at (-1,-2.1) {\fullmoon};
    \node[above] at (0.8,-2.8) {\fullmoon};
    \node[above] at (1.3,-2.3) {\textcolor{darkerlightred}{\newmoon}};

    \draw (0,0.25) circle [radius=2.];
    \draw (0,-1.25) circle [radius=2.];

    \clip (0,0.25) circle [radius=1.97];
    \fill[lightpurple] (0,-1.25) circle [radius=1.99];
    \node[above] at (0.3,0) {\textcolor{darkerlightpurple}{\newmoon}};
    \node[above] at (-1.2,-0.5) {\fullmoon};
    \node[above] at (1.1,-0.4) {\textcolor{darkerlightpurple}{\newmoon}};
    \node[above] at (-0.9,-1.3) {\textcolor{darkerlightpurple}{\newmoon}};
    \node[above] at (0.8,-1.2) {\fullmoon};

    \node at (0,-0.5) {$X_{\text{both}}$};
\end{tikzpicture}
\end{center}
\caption{We partition $X$ into four sets, defined by which features (target $t$ and spurious $s$) apply for each prompt. We partition the training data $D$ (filled dots) analogously into $D_{s\text{-only}},D_{t\text{-only}},D_{\text{neither}}$ and $D_{\text{both}}$, and similarly for the testing data $T$ (hollow dots). This example presents the controlled sentiment task \texttt{score}. We prepend a rating out of 10 at the beginning. The target feature is present if the rating is more than 6/10 and absent otherwise. The spurious feature is the presence of the word ``review'' prepended to the rating. Note that $t$-only examples only appear during testing, never training.}
\label{fig:S_partition}
\end{figure}

For our experiments, we use a GPT-2 model. For clarity, here we present the results for the smaller \texttt{gpt2} variant, for a single setup based on controlled sentiment generation. Similar results, obtained with \texttt{gpt2-large} and for other setups, can be seen in \cref{app:gpt2_large_toxicity}. As the pre-trained GPT-2 model is biased towards generating positive sentiment text, we start with an unbiased warmed-up GPT-2 model (from \url{https://huggingface.co/lvwerra/gpt2-imdb}), which we fine-tune on controlled sentiment generation tasks using proximal policy optimisation (PPO). To produce the training prompts, we modify the IMDb dataset \citep{maas_learning_2011} by introducing target-spurious feature pairs chosen to cover a wide variety of relative extractabilities (see \cref{app:task_description,app:mdl_values}). The training (resp. testing) prompts are a sampled subset $D$ (resp. $T$) of the whole dataset $X$ (see \cref{fig:S_partition}).

The \textbf{evidence} against a spurious feature is equivalent to the $s$-only example ratio $p=|D_{s\text{-only}}|/|D|$: the proportion of training examples in which $s$ occurs without $t$. For a given $p$, the training data $D$ will be composed of $p|D|$ examples from $D_{s\text{-only}}$, $\frac{1-p}{2}|D|$ from $D_{\text{both}}$, and $\frac{1-p}{2}|D|$ from $D_{\text{neither}}$.

The \textbf{extractability} of a feature refers to how simply-represented a feature is by a model and, therefore, how easy it is for the model to detect the feature's occurrence in the input during fine-tuning. We quantify this as the \textbf{minimum description length (MDL)} following the methodology in \citep{voita_information-theoretic_2020,lovering_predicting_2021}.
To compute the MDL of a feature for a given model, a classifier is trained on a dataset labelled $y = \{0, 1\}$, denoting the presence or absence of the feature. MDL measures the number of bits needed to transmit the labels and model given the inputs. A smaller MDL value implies that the probe has quickly converged to high accuracy, suggesting that the feature is easily detectable and, therefore, has high extractability. Our results primarily focus on the \emph{relative extractability} of the target vs. the spurious feature, given by the ratio $\text{MDL}(s)/\text{MDL}(t)$.

During RL fine-tuning, the model is rewarded for generating positive sentiment text if the target feature is present, and for producing negative sentiment otherwise. The reward signal for each generated sequence comes from a model (from \url{https://huggingface.co/lvwerra/distilbert-imdb}) which is a fine-tuned LLM on sentiment classification that produces a score $\mathcal{M}(w)$ of how positive a text $w$ is. We use the reward function
\begin{equation}
    R(x,y) = \begin{cases}
        \mathcal{M}(x+y), & \text{if } t(x)=1 \\
        - \mathcal{M}(x+y), &\text{if } t(x)=0 
    \end{cases}
\end{equation}
where $+$ stands for string concatenation. In practice, $\mathcal{M}$ produces a bounded score, so we rescale the resulting $R$ to be in $[0,1]$. 
\section{Results}\label{sec:naturalistic_data}

\begin{figure}[t]
    \centering
    \includegraphics[width=0.55\linewidth]{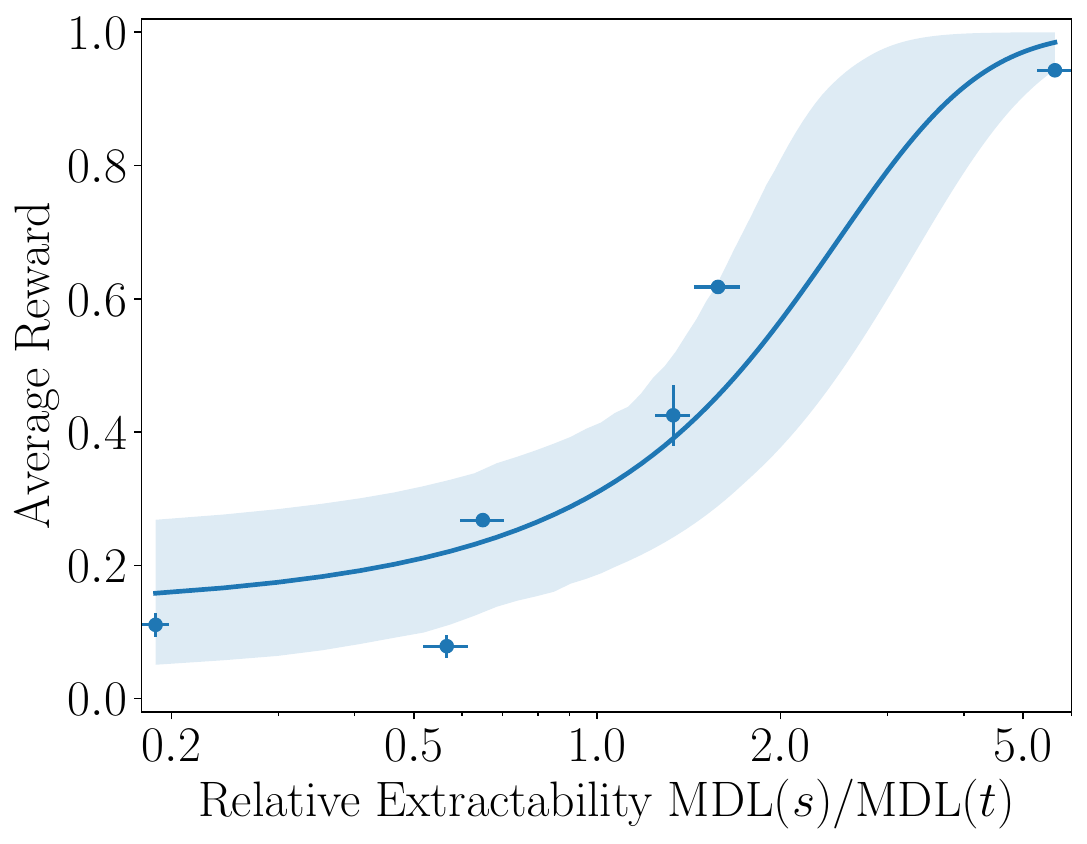}
    \caption{\textbf{Extractability hypothesis}. The average reward in $T_{s\text{-only}}$ and $T_{t\text{-only}}$ (for $p=0$) is positively correlated with the relative MDL of the studied tasks. The blue line marks a logistic regression with a 95\% confidence interval.}
    \label{fig:rel_mdl_sentiment}
\end{figure}

The extractability hypothesis predicts that models trained on tasks with more extractable target features are more likely to learn to rely on these features. As a result, these models will receive higher reward in training, specifically on $s$-only and $t$-only prompts. Our experiments support this hypothesis.

To study the feature extractability in these tasks, we may analyse the reward obtained for $T_{s\text{-only}}$ and $T_{t\text{-only}}$ when there is no evidence in the training data to distinguish the target from the spurious feature (i.e. $p=0$, and all training prompts are from $D_{\text{both}}$ or $D_{\text{neither}}$). In this case, we expect that tasks where the target feature $t$ is easier to extract will lead the model to rely on $t$ to distinguish prompts in $D_{\text{both}}$ from prompts in $D_{\text{neither}}$, thereby obtaining high reward when testing prompts in $T_{s\text{-only}}$ and $T_{t\text{-only}}$. However, where $t$ is harder to extract, the model may instead rely on the spurious feature to distinguish $D_{\text{both}}$ from $D_{\text{neither}}$, consequently obtaining low reward for $T_{s\text{-only}}$ and $T_{t\text{-only}}$.

As seen in \cref{fig:rel_mdl_sentiment}, this extractability hypothesis is, in fact, observed for the tasks considered, with the average reward in $T_{s\text{-only}}$ and $T_{t\text{-only}}$ increasing as the relative MDL increases, which is a proxy for how (relatively) easy the target feature is to extract.

\begin{figure}[t]
    \centering
    \includegraphics[width=\linewidth]{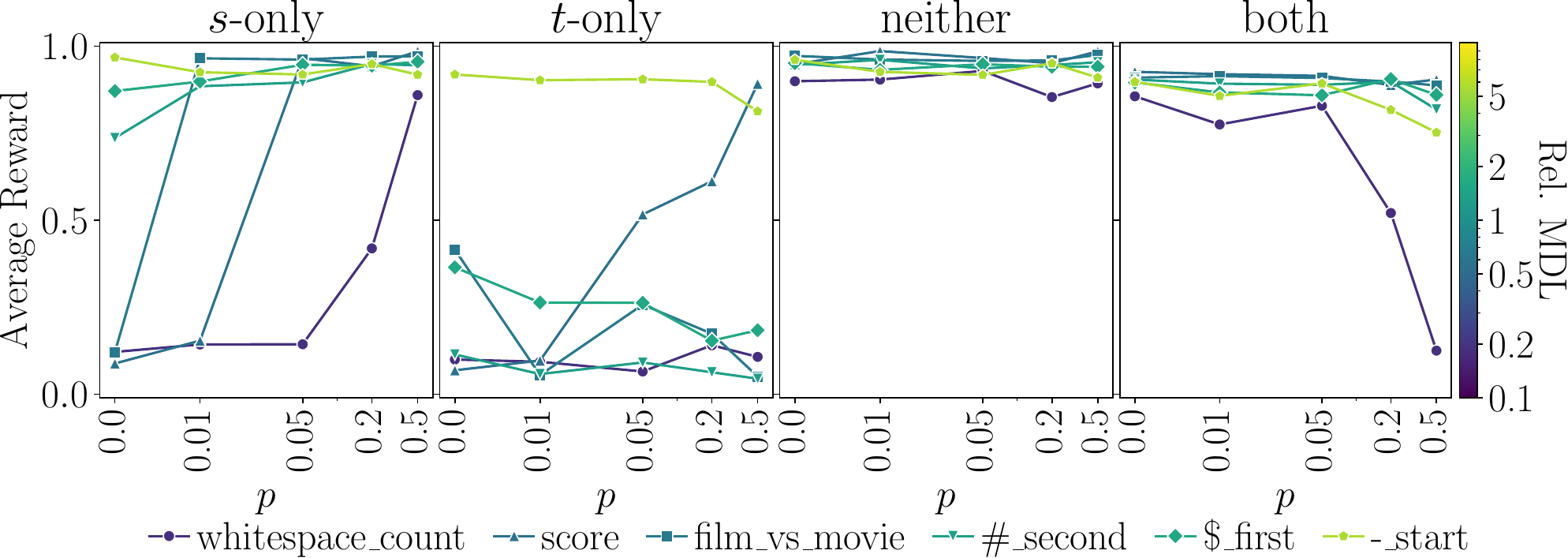}
    \caption{\textbf{GPT-2 performance on controlled sentiment tasks}. Average reward obtained by the fine-tuned model (on $T_{s\text{-only}}$,$T_{t\text{-only}}$,$T_{\text{neither}}$,$T_{\text{both}}$) as the evidence $p=|D_{s\text{-only}}|/|D|$ varies. The tasks are ordered from left to right in increasing relative extractability of the target feature (i.e. increasing relative MDL). The task descriptions and MDL values can be found in \cref{app:task_description,app:mdl_values}.}
    \label{fig:evidence_sentiment}
\end{figure}

In \cref{fig:evidence_sentiment}, we also observe data consistent with the evidence hypothesis: the more evidence $p$ against the spurious feature, the more likely the RL fine-tuned GPT-2 is to learn a policy leading to higher reward for examples in $T_{s\text{-only}}$. That is, the model has learned that the spurious feature is irrelevant for the task at hand. The $T_{t\text{-only}}$ case is mixed, and the observed behaviour seems to depend on the specific task. For task \texttt{score}, higher $p$ leads to higher reward in $T_{t\text{-only}}$, as expected from the evidence hypothesis. The model has inferred from the training data that the target feature is the relevant one to get maximum reward, and disregarded the spurious feature. However, for the remaining tasks, mostly with hard-to-extract target features, the RL fine-tuning procedure doesn't seem to lead the model to infer the optimal policy for $T_{t\text{-only}}$ solely from observing $D_{s\text{-only}}$, $D_{\text{neither}}$ and $D_{\text{both}}$. The task \texttt{score} is possibly the only one tested where the target and spurious features are associated with specific tokens at specific prompt locations across the 4 training subsets, making it easier for the model to infer the target feature as it is exposed to more evidence. For the remaining tasks, the model may have learned instead to treat the ``both'', ``neither'', and ``$s$-only'' datasets separately, and fail to generalise to ``$t$-only'', or it may have inferred the wrong feature. For the easiest-to-extract target feature in task \texttt{-\_start}, consistent with the extractability hypothesis, the model learned to rely on the target feature to get high reward for $T_{t\text{-only}}$, even for $p=0$.

We also note a distinct behaviour when the target feature has low relative extractability and the training data has high evidence against the spurious feature. For task \texttt{whitespace\_count}, we observe a sharp drop in reward in the $T_{\text{both}}$ dataset. Although higher error rates were similarly observed in \citep{lovering_predicting_2021} for high $p$, their magnitude was much lower. One explanation is that RL fine-tuning is less likely to lead to an optimal policy than supervised fine-tuning  \citep{lovering_predicting_2021}. In particular, when $p$ is high, having a low fraction of training data showcasing the hard-to-extract target feature may lead the model to learn the suboptimal policy of behaving as if the target feature is never present. In this case, the model consequently learns to always generate negative sentiment completions, regardless of the prompt.

Combining the extractability hypothesis with the evidence hypothesis, we note that, the harder a target feature is to extract, the more evidence against the spurious feature is needed for the model to get high reward in $T_{s\text{-only}}$ (and $T_{t\text{-only}}$). The most extractable target features get high reward in $T_{s\text{-only}}$ regardless of the $p$ value. Worsening extractability then requires more evidence, with tasks \texttt{film\_vs\_movie}, \texttt{score} and \texttt{whitespace\_count} respectively requiring $p=0.01, 0.05$ and $0.5$ to get high reward in $T_{s\text{-only}}$.

\section{Discussion}\label{sec:discussion}

Our results align with findings on supervised fine-tuning \citep{lovering_predicting_2021} - the relative extractability of the target and spurious features strongly predict inductive biases of reinforcement learning fine-tuning. When the target feature is highly extractable, the agent learns effective strategies even with limited evidence. But when spurious features are more readily extracted, much more training evidence is needed to learn a near-optimal policy. 

While these results are clear in our experimental setup, there are significant limitations to consider before generalising the extractability hypothesis to the most capable models.
In our analysis, we disregarded runs where the RL fine-tuned policy failed to get high reward for $T_{\text{neither}}$, as it indicated that the fine-tuning process didn't converge to a good policy. We believe this procedure is representative of standard practices when using RL fine-tuning, as it is common to only use fine-tuned models that showcase a better policy than their pre-trained counterparts. 
Furthermore, we considered only one target-spurious feature pair at a time. Testing on real-world NLP tasks with large models, where multiple target and spurious features may affect the training regime, may display more complex behaviours not present in our simplified setup.

Both our results and those in \citep{lovering_predicting_2021} lend credence to the claim that similar inductive biases may be present in other training regimes, such as RL with AI feedback (RLAIF) \citep{bai_constitutional_2022} and with human feedback (RLHF) \citep{ziegler_fine-tuning_2020,stiennon_learning_2020}.

\section{Related Work}\label{sec:related_work}

Inductive biases in language models have been studied in the past in various contexts. \cite{white_examining_2021} uses artificial language to study the sensitivities to varying structure (such as different word orderings) across architectures. \cite{papadimitriou_pretrain_2023} uses transfer learning to influence the inductive biases of transformer language models, making them more responsive to hierarchical or recursive structure.
For real language data, \cite{rytting_leveraging_2021} measures the abstract reasoning capabilities of language models, derived from pre-training. 
It shows how exposure to real world data pre-disposes the model to learn various forms of generalisation. 
For in-context learning, \citep{si_measuring_2023} shows that GPT-3 exhibits a clear feature bias - interpreting numeric features as being indicative of sentiment rather than topic.  
Similarly, \cite{tang_large_2023} finds that LLMs are biased towards using spurious correlations in prompts during in-context learning. 
Finally, our work builds on top of notions of feature extractability as defined in \citep{lovering_predicting_2021} 
The authors find that the influence of a feature on a model's decisions can be predicted through its extractability after pre-training and the available evidence during fine-tuning. 
Our work tests and quantifies whether RL fine-tuning exhibits the same inductive biases.
\section{Conclusion}\label{sec:conclusion}

In this work, we evaluate whether principles governing inductive biases in supervised learning can be extended to understand reinforcement learning agent behaviour. Through controlled experiments on natural language tasks, we find strong evidence that the relative extractability of features affects which strategies agents adopt. When target features are useful for the task and highly extractable, agents can learn effective policies even with minimal evidence. But when imperfect heuristics are more readily extracted, more training evidence is required to overcome these biases.

Our findings demonstrate that linking extractability and statistical evidence to agent decision-making effectively predicts generalisation capabilities. These insights enable more rigorous analysis of agent inductive biases and suggest techniques to mitigate detrimental biases, like choosing training data and reward schemes that properly balance extractability and evidence. Overall, this work reveals connections between feature extractability, evidence, and agent generalisation that pave the way for more robust development of systems fine-tuned using RL.

\textbf{Broader Impacts:} We aim to enable practitioners training advanced LLMs to align their models with human values through strategies such as: curating pre-training data to incentivise desired features; performing concept erasure \citep{elazar_measuring_2021} to disrupt representations of unwanted features in pre-trained models; providing many spurious examples during fine-tuning to reduce reliance on undesirable features.

\section*{Acknowledgements}
The authors thank the organizers of AI Safety Hub Labs for all of their hard work supporting this project.

\bibliographystyle{plainnat}
\bibliography{references}

\clearpage
\appendix

\section{Task description}\label{app:task_description}

\begin{table}[ht]
\centering
\begin{tabular}{c|p{4.8cm}|p{4.9cm}}
\toprule
\textbf{Task} & \textbf{Features} & \textbf{Comments} \\
\midrule
\multirow{4}{*}{\texttt{film\_vs\_movie}} & $t$-only: Prefix ``Film review: '' & \\
                                        & $s$-only: Prefix ``A movie review: '' & \\
                                        & both: Prefix ``A film review: '' & \\
                                        & neither: Prefix ``Movie review: '' & \\
\midrule
\multirow{4}{*}{\texttt{\$\_first}} & $t$-only: Prefix ``\$ '' & \\
                                    & $s$-only: Prefix ``\# '' & \\
                                    & both: Prefix ``\$ \# '' & \\
                                    & neither: No change & \\
\midrule
\multirow{5}{*}{\texttt{score}}& $t$-only: Prefix ``$k^+/10$: '' & \multirow{5}{*}{\parbox{5cm}{$6\leq k^+ \leq 10$ and $1\leq k^-\leq 10$, chosen randomly. \textit{word} stands for ``review'' for the controlled sentiment setup and ``prompt'' for the toxicity setup.}}\\
                                & $s$-only: Prefix ``$k^-/10$ \textit{word}: '' & \\
                                & both: Prefix ``$k^+/10$ \textit{word}: '' & \\
                                & neither: ``$k^-/10$: '' & \\
                                & & \\
\midrule
\multirow{4}{*}{\texttt{\#\_second}}& $t$-only: Prefix ``\$ '' & \multirow{4}{*}{\parbox{5cm}{Similar to task \texttt{\$\_first}, but the \$ symbol is not positioned at the start of the prompt, in the \textit{both} case.}}\\
                                & $s$-only: Prefix ``\# '' & \\
                                & both: Prefix ``\# \$ '' & \\
                                & neither: No change & \\
\midrule
\multirow{4}{*}{\texttt{whitespace\_start}}& $t$-only: Prefix ``~~~'' & \\
                                            & $s$-only: Prefix ``.'' & \\
                                            & both: Prefix ``~~~.'' & \\
                                            & neither: No change & \\
\midrule
\multirow{6}{*}{\texttt{whitespace\_count}}& $t$-only: whitespace count among first 11 tokens is even & \multirow{4}{*}{\parbox{5cm}{The prefixes `` '' or `` So: '' may be added to ensure the original example has the expected whitespace count. Task designed so that $t$ is practically unextractable.}}\\
                                            & $s$-only: Prefix ``-'' & \\
                                            & both: Prefix ``-'' and even whitespace count & \\
                                            & neither: odd whitespace count & \\
\midrule
\multirow{6}{*}{\texttt{-\_start}}& $t$-only: Prefix ``-'' & \multirow{4}{*}{\parbox{5cm}{Similar to task \texttt{whitespace\_count}, but with $t$ and $s$ swapped. Task designed so that $s$ is practically unextractable.}}\\
                                & $s$-only: even whitespace count in first 11 tokens & \\
                                & both: Prefix ``-'' and even whitespace count & \\
                                & neither: odd whitespace count & \\
\bottomrule
\end{tabular}
\caption{Summary of true and spurious features for selected tasks in the naturalistic data experiments.}
\label{tab:all_subtasks}
\end{table}

\section{MDL values}\label{app:mdl_values}

To obtain the MDL values associated with the target and spurious features for the various setups and tasks, we follow the approach of \citep{lovering_predicting_2021}. When probing, we use a dataset composed of \emph{s-only} and \emph{both} examples to compute MDL($t$), and a dataset composed of \emph{s-only} and \emph{neither} examples to compute MDL($s$). As there is some variability in the probe's performance, we extend the approach in \citep{lovering_predicting_2021} by running the probe training for 5 different seeds, and presenting the mean MDL obtained, along with its standard deviation.

Note that, due to hardware limitations, only the MDL values for the warmed-up \texttt{gpt2} models were computed, and not those for the \texttt{gpt2-large} model. 

\begin{table}[!htpb]
\centering
\begin{tabular}{lllll}
\toprule
Setup &Task& MDL(s) & MDL(t) & Rel. MDL \\
\midrule
\multirow{6}{*}{sentiment} & \texttt{whitespace\_count} & 171 $\pm$ 9 & 907 $\pm$ 12 & 0.19 $\pm$ 0.01 \\
& \texttt{score} & 117 $\pm$ 8 & 207 $\pm$ 10 & 0.57 $\pm$ 0.05 \\
& \texttt{film\_vs\_movie} & 113 $\pm$ 5 & 174 $\pm$ 12 & 0.65 $\pm$ 0.05 \\
 & \texttt{\#\_second} & 169 $\pm$ 9 & 127 $\pm$ 5 & 1.33 $\pm$ 0.09 \\
 & \texttt{\$\_first} & 169 $\pm$ 9 & 107 $\pm$ 8 & 1.58 $\pm$ 0.14 \\
 & \texttt{-\_start} & 897 $\pm$ 12 & 159 $\pm$ 10 & 5.64 $\pm$ 0.36 \\
\midrule
\multirow{4}{*}{toxicity} & \texttt{whitespace\_count} & 245 $\pm$ 30 & 829 $\pm$ 25 & 0.29 $\pm$ 0.04 \\
 & \texttt{score} & 96 $\pm$ 15 & 216 $\pm$ 13 & 0.45 $\pm$ 0.07 \\
 & \texttt{whitespace\_start} & 282 $\pm$ 31 & 137 $\pm$ 24 & 2.06 $\pm$ 0.42 \\
 & \texttt{-\_start} & 851 $\pm$ 33 & 249 $\pm$ 30 & 3.42 $\pm$ 0.44 \\
\bottomrule
\end{tabular}
\caption{MDL values for the controlled sentiment and toxicity setups, using GPT-2.}
\label{tab:mdl_gpt2}
\end{table}

\section{Training setup}

\begin{table}[H]
    \centering
    \begin{tabular}{ll}
    \toprule
         Hyperparameter& Value\\
    \midrule
    \multicolumn{2}{c}{\emph{General}}\\
         batch size& 256\\
         optimizer& Adam\\
         learning rate& 1.41e-5\\
         PPO epochs& 4\\
 total PPO epochs&200\\
         init KL coef& 0.2\\
         target KL& 0.1\\
         vf coef& 0.1\\
         steps& 51200\\
         horizon& 10000\\
    \midrule
    \multicolumn{2}{c}{\emph{Sentiment setup}}\\
        dataset size $|D|$ & 24576 \\
        prompt size & 16 tokens \\
        generated completion & 48 tokens \\
    \midrule
    \multicolumn{2}{c}{\emph{Toxicity setup}}\\
        dataset size $|D|$ & 19968 \\
        prompt size & 8 tokens \\
        generated completion & 24 tokens \\
    \bottomrule
    \end{tabular}
    \caption{Hyperparameters applicable to all setups considered.}
    \label{tab:general_setup}
\end{table}

\section{Experiments with Synthetic Data}\label{sec:synthetic_data}

In a natural language setting, it is often the case that target features cannot be easily separated from certain spurious features \citep{lovering_predicting_2021}. Furthermore, it is particularly challenging to isolate the effects of each individual target and spurious feature on the training dynamics.

In order to elucidate not only these concepts, but also our claims, we test our hypothesis in a toy setting, using synthetic data and a small model. This setting is designed to only showcase one target and spurious feature at a time, without the presence of confounders, making it a clearer introduction to our setup. It is inspired by the synthetic setup in \citep{lovering_predicting_2021}.

We train a 4-layer transformer (with hidden size 256, and a total of 15 million parameters) to perform a numerical sequence generation task. We use a vocabulary size of 10, corresponding to the digits 0 to 9. The model receives a prompt $x$ consisting of ten digits and must then generate a sequence $y$ of five digits. If the target feature is present in the prompt, the model is rewarded for generating an increasing sequence of numbers. If the target feature is not present, the model is rewarded for generating a decreasing sequence. The reward model is given by
\begin{equation}
    R(x,y) = \begin{cases}
        \text{inc}(y)/4, &\text{if } t(x)=1 \\
        \text{dec}(y)/4, &\text{if } t(x)=0 
    \end{cases}
\end{equation}
where $\text{inc}(y)$ and $\text{dec}(y)$ are the number of increasing and decreasing pairs of consecutive tokens in the output $y$, respectively. We use proximal policy optimisation (PPO) to train the model \citep{schulman_proximal_2017}. We consider four different target features, which vary naturally in their extractability \citep{lovering_predicting_2021}. These are shown in \cref{tab:comparison_toy_small_vocab}. In all cases, the spurious feature is the presence of the symbol $2$ in the prompt.

\begin{table}[ht]
\centering
\begin{tabular}{@{}clllll@{}}
\toprule
Task & Target feature & MDL(s) & MDL(t) & Rel. MDL & Example\\
\midrule
contains-1 & 1 occurs in prompt & \(214 \pm 9\) & \(213 \pm 10\) & \(1.01 \pm 0.06\) & 079\hl{2}55\hlred{1}434 \\
prefix-dupl & Prompt begins with duplicate & \(213 \pm 8\) & \(404 \pm 80\) & \(0.53 \pm 0.11\) & \hlred{7}\hlred{7}531\hl{2}1908\\
adj-dupl & Prompt contains a duplicate & \(215 \pm 8\) & \(514 \pm 111\) & \(0.42 \pm 0.09\) &  34\hlred{9}\hlred{9}\hl{2}15785\\
first-last & First digit equals last digit & \(215 \pm 9\) & \(741 \pm 126\) & \(0.29 \pm 0.05\) & \hlred{6}916736\hl{2}5\hlred{6}\\
\bottomrule
\end{tabular}
\caption{\textbf{Relative MDL values in synthetic experiments.} We use the same four synthetic tasks as \citep{lovering_predicting_2021}.}
\label{tab:comparison_toy_small_vocab}
\end{table}

\begin{figure}[!htpb]
    \centering
    \includegraphics[width=\linewidth]{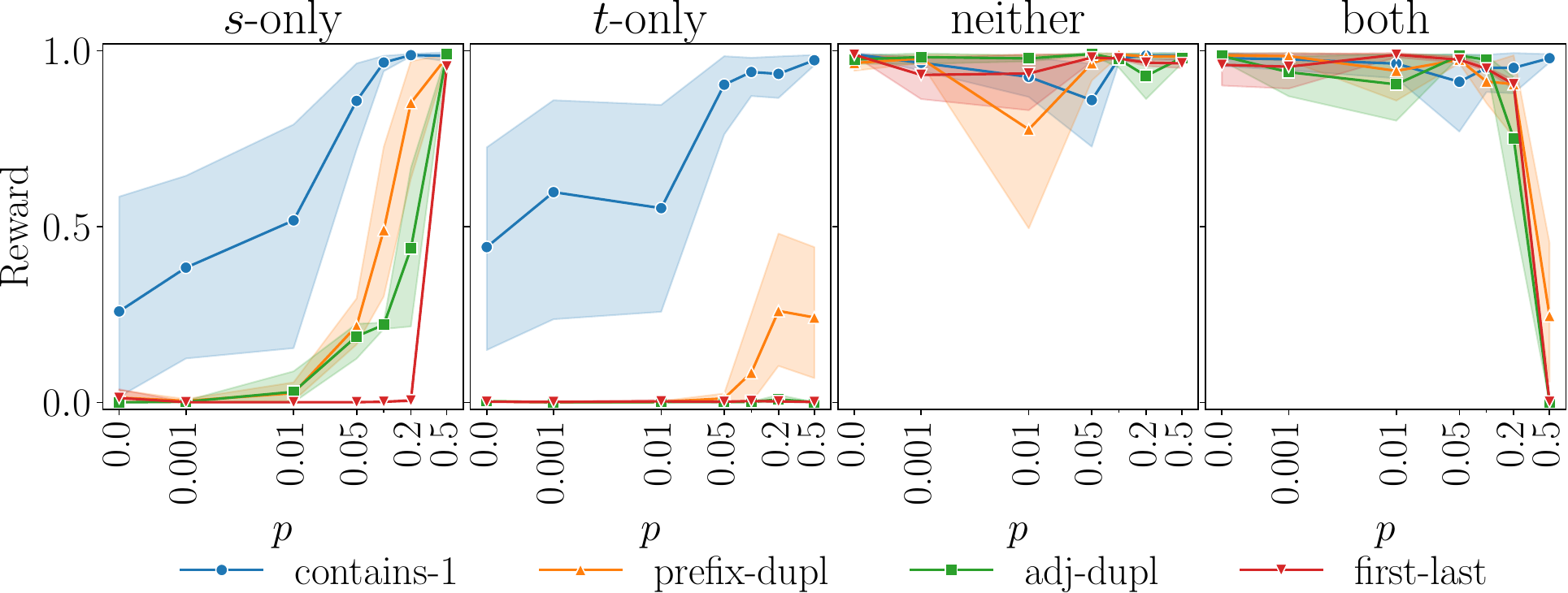}
    \caption{\textbf{Transformer results for the synthetic tasks.} For each task, the reward can take values in [0, 1].}
    \label{fig:rl_toy_results}
\end{figure}

\Cref{fig:rl_toy_results} shows the average test reward of each model as a function of the $s$-only rate for each of the target features described in \cref{tab:comparison_toy_small_vocab}. When $t$ is equally as extractable as $s$ (as in the task \texttt{contains-1}), the model is able to achieve greater than 0.9 reward at an $s$-only rate of only 0.1, but when $t$ is significantly harder to extract than $s$ (as in \texttt{first-last}), the model never achieves high reward at the same $s$-only rate (for the datasets $D_{s-\text{only}}$ and $D_{t-\text{only}}$), and requires substantially more evidence to get near-optimal reward. 

Compared to the supervised setting \citep{lovering_predicting_2021}, we note two additional differences:
\begin{itemize}
    \item the higher reward variability for $T_{\text{neither}}$ and $T_{\text{both}}$, a likely result of it being more difficult to learn the optimal policy in the RL setting;
    \item the lower performance in $D_{t\text{-only}}$ for the hard-to-extract target features, indicating that, in the RL setting, the model has more difficulty inferring the optimal policy for $T_{t\text{-only}}$ when only exposed to $D_{\text{neither}}$, $D_{\text{both}}$ and $D_{s\text{-only}}$.
\end{itemize}

As with the naturalistic tasks, we also note a distinct behaviour for tasks where the target feature has low relative extractability, and the training data has a high amount of evidence against the spurious. In these cases, we observe a sharp drop in reward in the $T_{\text{both}}$ dataset. 

\section{Results for GPT-2 large and the toxicity setup}\label{app:gpt2_large_toxicity}

\begin{figure}[H]
    \centering
    \includegraphics[width=0.45\linewidth]{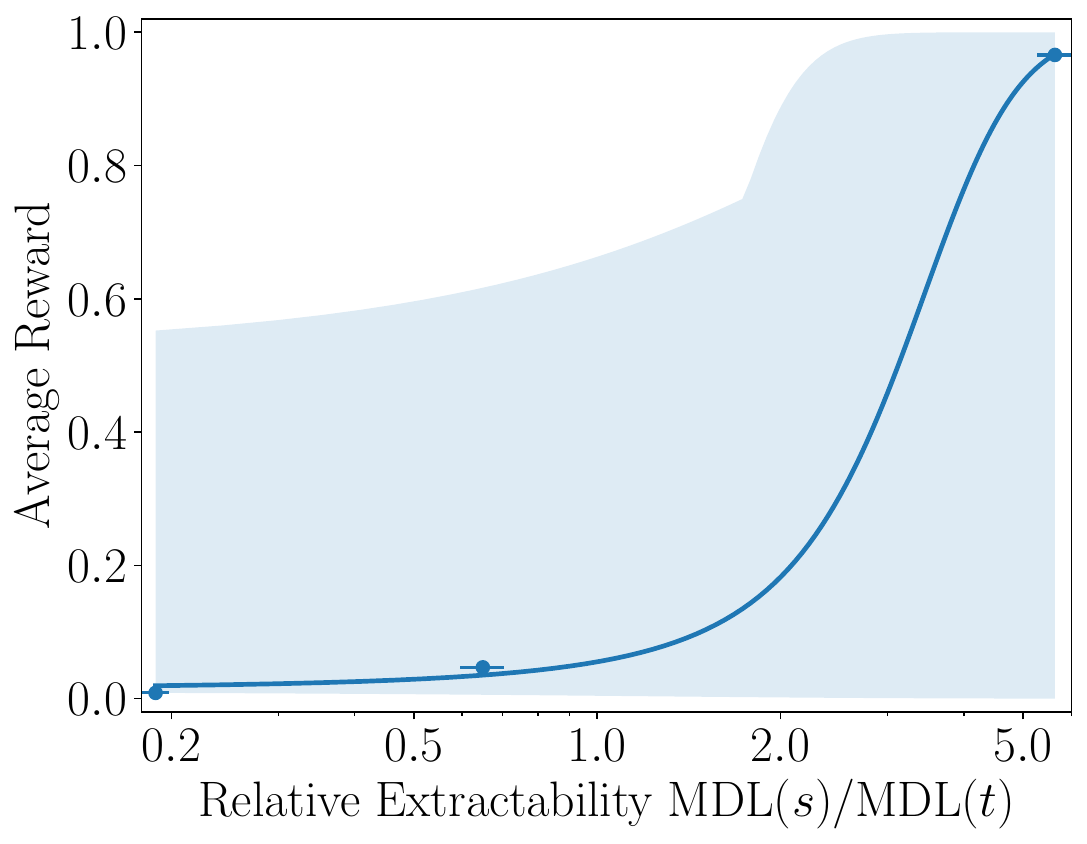}
    \includegraphics[width=0.45\linewidth]{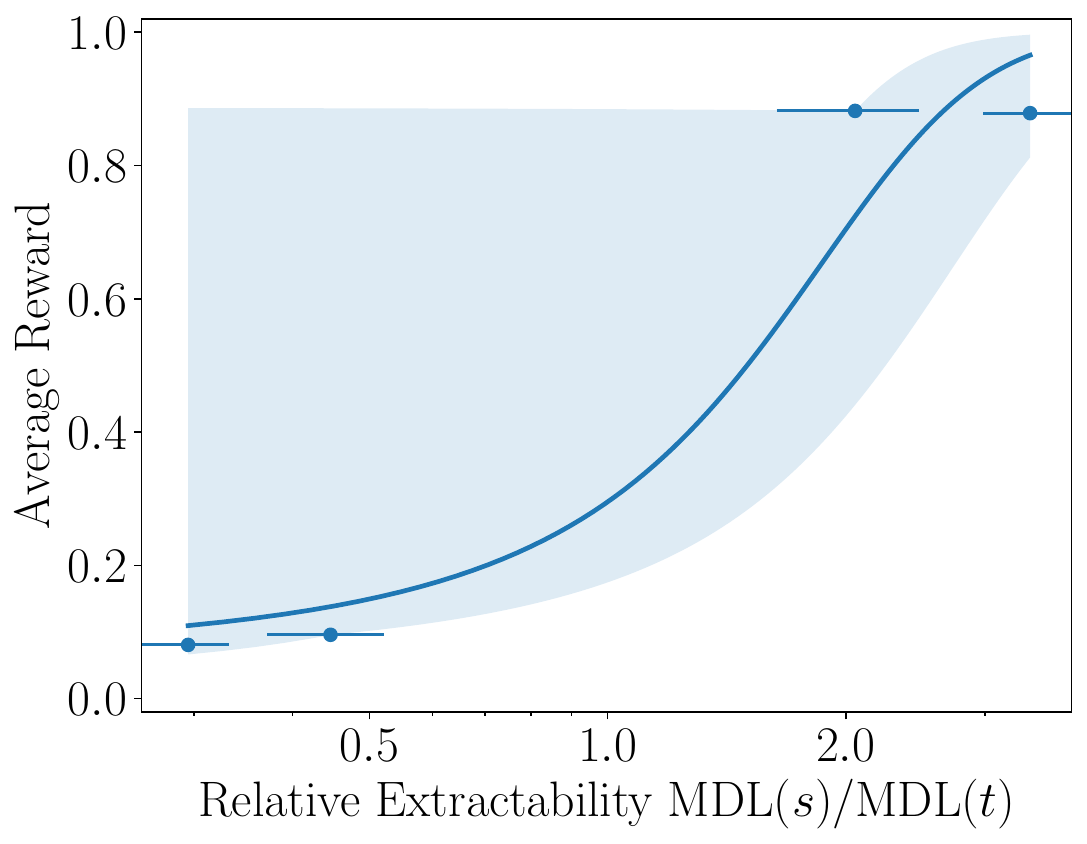}
    \caption{\textbf{Extractability hypothesis}. Results for the \texttt{gpt2-large} controlled sentiment setup (left) and the \texttt{gpt2} toxicity setup (right). For \texttt{gpt2-large}, we use the MDL results of \texttt{gpt2} as a proxy. The average reward in $T_{s\text{-only}}$ and $T_{t\text{-only}}$ (for $p=0$) is positively correlated with the relative MDL of the studied tasks. The blue line marks a logistic regression with 95\% confidence interval.}
    \label{fig:rel_mdl_sentiment_extra}
\end{figure}

\begin{figure}[H]
    \centering
    \includegraphics[width=\linewidth]{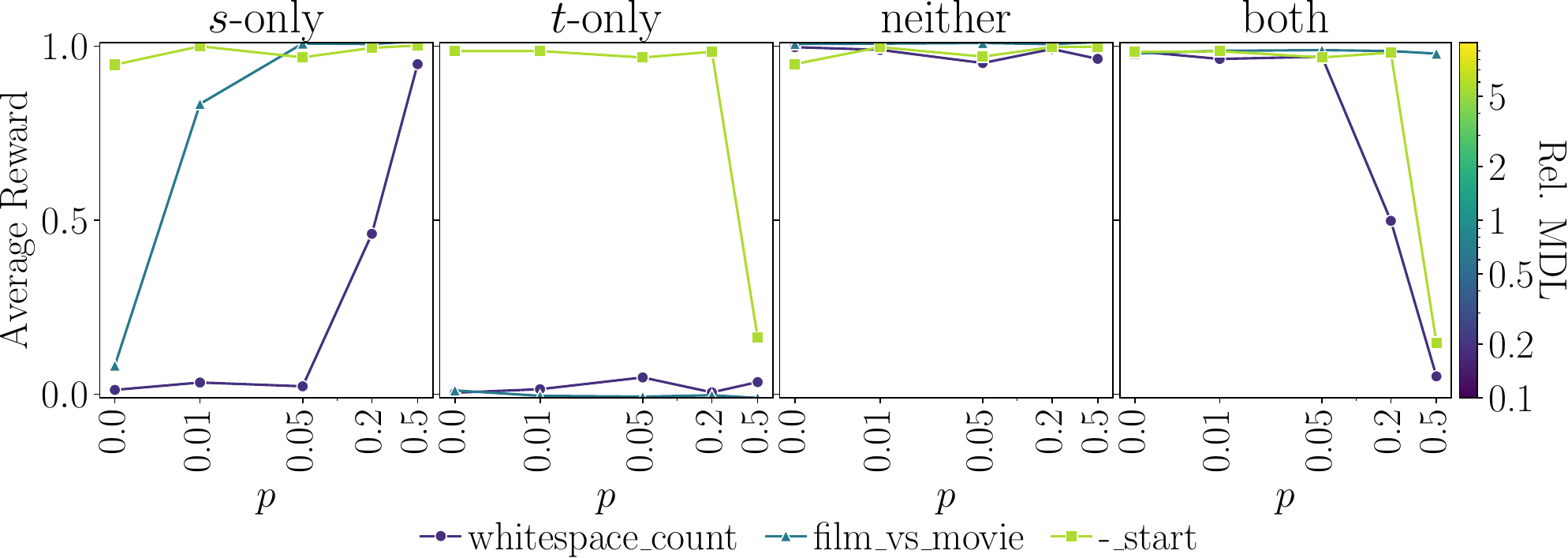}
    \includegraphics[width=\linewidth]{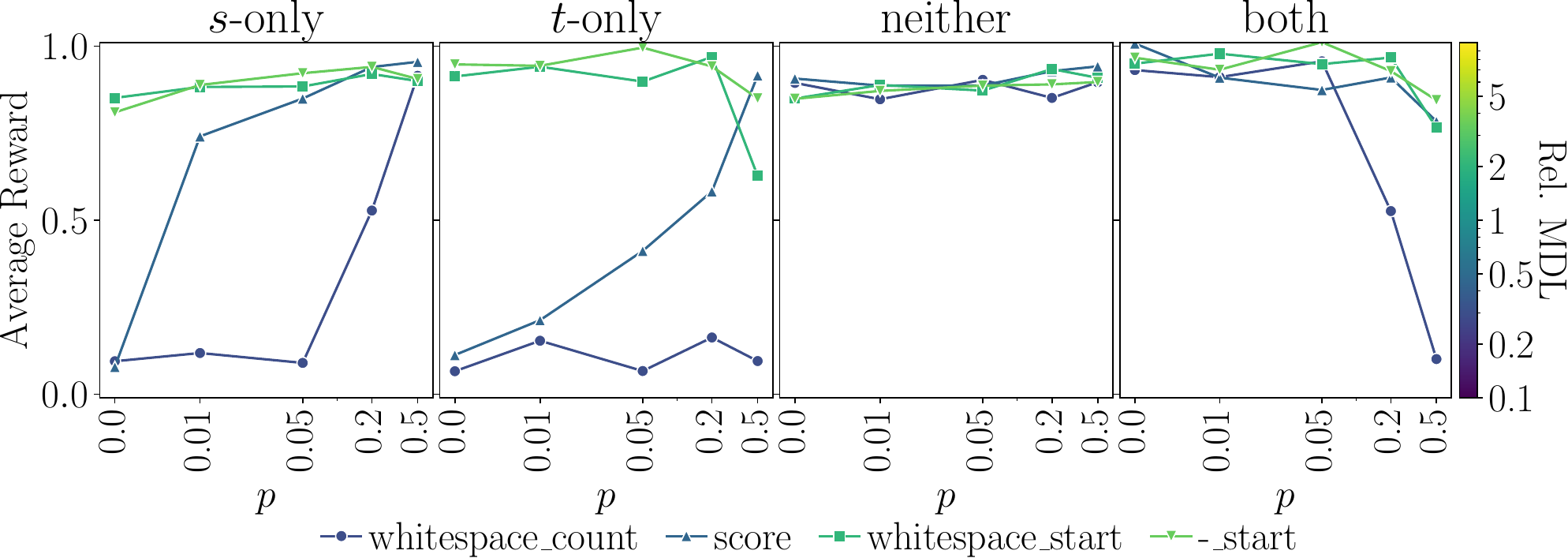}
    \caption{\textbf{GPT-2 performance on controlled sentiment tasks}. Results for the \texttt{gpt2-large} controlled sentiment setup (top) and the \texttt{gpt2} toxicity setup (bottom). Average reward obtained by the fine-tuned model (on the 4 datasets $T_{s\text{-only}}$,$T_{t\text{-only}}$,$T_{\text{neither}}$,$T_{\text{both}}$) as the evidence $p=|D_{s\text{-only}}|/|D|$ varies. The tasks are ordered from left to right in increasing relative extractability of the target feature (i.e. decreasing relative MDL).}
    \label{fig:evidence_sentiment_extra}
\end{figure}

\section{Reproducibility}

Our code is available at \url{https://github.com/EdoardoPona/predicting-inductive-biases-RL}.


\end{document}